\documentclass[12pt,journal,compsoc,onecolumn]{IEEEtran}
\usepackage{amsmath,epsfig}
\usepackage{amssymb}
\usepackage{xcolor}
\usepackage[noadjust]{cite}

\def\x{{\mathbf x}}
\def\L{{\cal L}}

\title{Reinforcement-based frugal learning for satellite image change detection}
\author{ Sebastien Deschamps$^{1,2}$ \ \ \ \ \ \  \ \ \ \ \ \  \ \ \ \ \ \   \ \ \ \ \ \ Hichem Sahbi$^1$ \\  $ $ \\
$^1$Sorbonne University, UPMC, CNRS, LIP6, France \\  $^2$Theresis Thales, France}
\begin{document}

\maketitle

\begin{abstract}
In this paper, we  introduce a novel interactive satellite image change detection algorithm based on active learning.  The proposed approach is  iterative and asks the user (oracle) questions about  the targeted changes and according to the oracle's responses updates change detections.  We consider a probabilistic framework which assigns to each unlabeled sample a relevance measure modeling how critical is that sample when training  change detection functions.   These relevance measures are obtained by minimizing an objective function mixing diversity, representativity and uncertainty.  These criteria  when combined allow exploring different data modes and also refining change detections.  To further explore the potential of this objective function, we consider a reinforcement learning approach that  finds the best combination of diversity, representativity and uncertainty,  through  active learning iterations, leading to better generalization as corroborated through  experiments in interactive satellite image change detection. 
\end{abstract}

\section{Introduction}
\label{sec:intro}

Satellite image change detection consists in finding  occurrences of targeted (relevant) changes into a scene at a given instant w.r.t. the same scene acquired earlier  \cite{ref7,ref9,ref11,ref13}.  This includes  appearance or disappearance of  visual entities  such as  infrastructure destruction after natural hazards (earthquakes, tornadoes, etc.) \cite{ref4,ref5}.  This task is  very challenging as relevant changes are eclectic and satellite images are   subject to multiple sources of irrelevant changes including illumination,  artefacts, clouds, etc.  Existing solutions either remove  irrelevant variations in satellite images by correcting their radiometric effects \cite{ref14,ref15,ref17,ref20} or consider them as a part of appearance modeling \cite{ref21,ref25,ref26,ref27,ref28}.  The latter consists in designing statistical or machine inference models that  learn how to discriminate between relevant and irrelevant changes.  Training these models requires enough labeled data covering all the sources of variability due to both the positive and negative classes.  However,  beside the scarceness of labeled data, the relevance of changes could be subjective and may vary from one user to another, and this makes the task of automatic change detection highly challenging. \\

\indent Existing machine learning approaches that mitigate the scarceness of labeled data include few shot, self-supervised and active learning \cite{reff45,refff2,reff2,reff16,reff15,reff53,reff12,reff74,reff58,reff1}.  Among these methods,  active learning is particularly interesting and allows modeling the user's subjectivity (about  targeted changes) more accurately.  Active learning solutions are interactive approaches that show the most critical unlabeled data (a.k.a. displays) to the user/oracle, and ask the latter about the relevance of changes prior to update  change detections \cite{refff33333}.  Display section strategies usually rely on multiple criteria including diversity, representativity and uncertainty  \cite{reff13}.   Diversity allows exploring different  modes of the unlabeled data, representativity seeks to select prototypical samples in those modes while uncertainty allows displaying the most ambiguous data that ultimately refine change detections.  However, knowing a priori which sequence of display selection  strategies (diversity, representativity and uncertainty) to apply  through all the iterations of active learning is highly combinatorial.  Besides,  under the regime of frugal learning,  labeled validation sets are scarce in order to make the optimization of these  strategies  statistically meaningful. \\

\indent In this paper, we devise a novel change detection  algorithm that asks the oracle  the most informative questions about  targeted changes and according to the oracle's responses updates change detections.  The proposed solution is probabilistic and assigns to each unlabeled sample a relevance measure which captures how critical is that sample when learning changes.  These relevance measures are obtained as the optimum of  an objective function that  mixes diversity,  representativity and ambiguity criteria.  In order to tackle the combinatorial aspect of these  criteria, we further rely on  reinforcement learning (RL) which finds the ``optimal''  sequence of actions (diversity,  representativity and ambiguity as well as their possible combination) that ultimately leads to  high generalization.  Experiments conducted on the challenging task of interactive satellite image change detection show the superiority and the outperformance of the proposed RL-based   approach w.r.t.  related work.

\def\X{{\cal X}}
\def\L{\cal L}  
\def\U{\cal X}  
\def\D{{\cal D}}
 \def\Y{{\cal Y}}
\def\x{{\bf  x}}
\def\C{{\bf C}}
\def\F{{\bf F}}
\def\DD{{\bf D}}
\def\tr{{\bf tr}}
\def\1{{\bf 1}}
\def\S{{\cal S}}
\def\A{{\cal A}}
\def\T{{T}}
\def\R{{\cal R}}
\def\O{{\textrm{oracle}}}
 \def\diag{{\textrm{diag}}}

\section{Proposed model}
\label{sec:proposed}

Let ${\cal I}_r=\{p_1,\dots,p_n\}$, ${\cal I}_t=\{q_1,\dots,q_n\}$  denote two registered satellite images taken at two different time-stamps $t_0$, $t_1$ respectively, and let $\X=\{\x_1,\dots,\x_n\}$ be a set of aligned patch pairs with $\x_i=(p_i, q_i) \in {\cal I}_r \times {\cal I}_t$.  Considering the labels of $\X$ initially unknown, our goal is to design a classifier $g(.)$ by interactively labeling  a very {\it small} fraction of $\X$ (as change / no-change), and  training the parameters of $g$. This interactive labeling and training  is known as active learning.\\ 
\noindent Let  $\D_t$ be a {\it display} (defined as a subset of $\X$) shown to an oracle\footnote{The oracle is defined as an expert annotator providing labels (changes / no-changes) for any given subset of images.} at any iteration $t$ of active learning, and let $\Y_t$ be the underlying labels.   The initial display $\D_t$ (with $t=0$) is  uniformly sampled at random, and used to train  the subsequent classifiers  by repeating the following steps till reaching high generalization  or exhausting a  labeling budget:  \\
\noindent i)  Get the labels of  $\D_t$ as $\Y_t \leftarrow \O(\D_t)$.\\
\noindent ii) Train $g_t(.)$ using $\bigcup_{\tau=1}^t (\D_\tau,\Y_\tau)$ where the subscript in  $g_t(.)$ refers to the decision function at  iteration $t$.  In  this paper, support vector machines built on top of  convolutional features are used.  \\
\noindent iii) Select the next display $\D \subset {\X}-\bigcup_{\tau=1}^t \D_\tau$ that possibly increases the generalization performances of the subsequent classifier $g_{t+1}(.)$. As the labels of  $\D$ are unknown,  one cannot combinatorially sample all the possible subsets  $\D$,  train the associated classifiers, and select the best display.  Alternative display selection strategies (a.k.a display models) are usually related to active learning and seek to find the most representative display that eventually yields optimal decision functions \cite{refff1}. \\ 
In what follows, we introduce our main contribution: a novel display model which allows selecting the most representative samples to label by an oracle and ultimately  lead to high generalization performances, in satellite image change detection, as corroborated  later in experiments. 
\subsection{Display model}
\indent We consider a probabilistic framework  which assigns for each sample $\x_i \in \X$ a membership degree $\mu_i$ that measures the probability of $\x_i$ belonging to the next display $\D_{t+1}$; consequently, $\D_{t+1}$ will correspond to the unlabeled data in $\{\x_i\}_i \subset \X$ with the highest memberships $\{\mu_i\}_i$. Considering $\mu \in \mathbb{R}^n$ (with $n= |{\X}|$) as a vector of these memberships  $\{\mu_i\}_i$, we propose to find $\mu $ as the minimum of the following constrained optimization problem 
{\begin{equation}\label{eq0}
  \begin{array}{l}
    \displaystyle    \min_{\mu \geq 0, \|\mu\|_1=1}  \eta \ \tr\big(\diag(\mu' [\C \circ \DD])\big) + \alpha  \ [\C' \mu]' \log [\C' \mu] \\
             \ \ \ \ \ \ \  \ \ \ \ \ \ \ \ \ \ \    \ \   + \beta \  \tr\big(\diag(\mu' [\F \circ \log \F]) \big) +  \mu' \log \mu, 
 \end{array}  
\end{equation}}
\noindent here $\circ$, $'$ are respectively the Hadamard product and the matrix transpose, $\|.\|_1$ is the $\ell_1$ norm, $\log$ is applied entry-wise, and $\diag$ maps a vector to a diagonal matrix.  In the above objective function (i) $\DD \in \mathbb{R}^{n \times K}$ and $\DD_{ik}=d_{ik}^2$ is the euclidean distance between $\x_i$ and $k^{\textrm{th}}$ cluster centroid of a partition of $\X$  obtained with K-means clustering, (ii) $\C \in \mathbb{R}^{n \times K}$ is the indicator matrix with each entry  $\C_{ik}=1$ iff $\x_i$ belongs to the $k^{\textrm{th}}$ cluster ($0$ otherwise),  and (iii) $\F \in \mathbb{R}^{n \times 2}$ is a scoring matrix with $(\F_{i1},\F_{i2})=(\hat{g}_t(\x_i),1-\hat{g}_t(\x_i))$ and $\hat{g}_t \in [0,1]$ being a normalized version of $g_t$. The first term of this objective function (rewritten as $\sum_i \sum_k  1_{\{\x_i \in h_k \}} \mu_i d_{ik}^2$) measures the {\it representativity} of the selected samples in $\D$; in other words, it captures how close is each $\x_i$ w.r.t. the centroid of its cluster, so this term reaches its smallest value when all the selected samples coincide with these centroids. The second term (rewritten as $\sum_{k}   [\sum_{i=1}^n 1_{\{\x_i \in h_k \}} \mu_i] \log [\sum_{i=1}^n 1_{\{\x_i \in h_k \}} \mu_i]$) measures the {\it diversity} of the selected samples as the entropy of the probability distribution of the underlying clusters; this measure is minimized when the selected samples belong to different clusters and vice-versa. The third criterion (equivalent to $\sum_i \sum_c^{nc} \mu_i \F_{ic} \log \F_{ic} $) captures the {\it ambiguity} in $\D$ measured as the entropy of the scoring function; this term reaches its smallest value when  data are evenly scored w.r.t.  different categories. Finally, the fourth term is related to the {\it cardinality} of $\D$, measured by the entropy of the distribution $\mu$; this term also acts as a regularizer.  Considering $\1_{nc}$, $\1_{K}$ as vectors of $nc$ and $K$ ones respectively (with $nc=2$ in practice),  one may show that the solution of Eq.~\ref{eq0} is given as $\mu^{(\tau+1)} :=\displaystyle \hat{\mu}^{(\tau+1)}/\|\hat{\mu}^{(\tau+1)}\|_1$, with  $\hat{\mu}^{(\tau+1)}$ being
\begin{equation}\label{eq2}
  \exp\bigg(-\big[\eta(\DD\circ \C)\1_K + \alpha \C (\log[\C' {\mu}^{(\tau)}]+\1_K)+\beta (\F \ \circ \  \log \F)\1_{nc}\big] \bigg). 
\end{equation}
\noindent As shown subsequently, the setting of the hyper-parameters $\alpha, \beta, \eta$  is crucial for the success of the  display model.  For instance, putting  emphasis on diversity (i.e.,  $\alpha \neq 0$) results into  exploration of class modes while a   focus on ambiguity (i.e.,   $\beta \neq 0$) locally refines the trained decision functions.  A suitable balance between exploration  and local refinement of the learned decision functions should be achieved by selecting the best configuration of these hyper-parameters.  Nevertheless,  since labeling is sparingly achieved by the oracle,  no sufficiently large validation sets could be made  available beforehand to accurately set these hyper-parameters. 
\subsection{RL-based display model}\label{rlbased}
\indent Let $\Lambda_\alpha$,  $\Lambda_\beta$, $\Lambda_\eta$ denote the parameter spaces associated to $\alpha$,  $\beta$, $ \eta$ respectively,  and let $\Lambda$  be the underlying   Cartesian product.  For any instance  $\lambda \in \Lambda$ (at  a given iteration $t+1$),  one may obtain a display (now rewritten as  $\D^\lambda_{t+1}$) by solving   Eq.~\ref{eq0}.  In order to find  the best configuration $\lambda^*$ that yields an ``optimal'' display,  we model hyper-parameter selection as a Markov Decision Process (MDP). An MDP based RL corresponds to a tuple $\langle \S,\A,R,\T,\delta\rangle $ with $\S$ being a state set, $\A$ an action set,  $R: \S \times \A \mapsto \mathbb{R}$ an immediate  reward function,  $T: \S\times \A \mapsto \S$ a transition function and $\delta$ a discount factor \cite{refff333334}.   RL consists in running a sequence of actions from $\A$ with the goal of maximizing an expected discounted reward by following a stochastic policy, $\pi: \S \mapsto \A$;  this leads to the true state-action value as  
\begin{equation}
Q(s,a) = E_\pi \left[  \sum_{k=0}^\infty  \delta^k {r}_k | S_0,=s,  A_0=a \right],
\end{equation} 
here  $E_\pi$ denotes the expectation w.r.t.  $\pi$,   ${r}_k$ is the immediate reward at the $k^{\textrm{th}}$ step of RL,   $S_0$ an initial state,  $A_0$ an initial action  and $\delta \in [0,1]$ is a discount factor that balances between immediate and future rewards. The goal of the optimal policy is to select actions that maximize the  discounted cumulative reward; i.e., $\pi_{*}(s) \leftarrow \arg\max_{a} Q(s,a)$.  One of the most used methods to solve this type of RL problems is Q-learning \cite{refff333335}, which directly estimates the optimal value function and obeys the fundamental identity, the Bellman equation 
\begin{equation}
Q_*(s,a) = E_\pi \left[ {R}(s,a) + \delta \max_{a'} Q_*(s',a') | S_0=s,A_0=a \right],
\end{equation} 
with $s'=T(s,a)$ and ${ R}(s,a)$ is again the immediate reward.    We consider in our hyper-parameter optimization, a stateless version,  so $Q(s,a)$ and $R(s,a)$ are rewritten simply as $Q(a)$, $R(a)$ respectively. In this configuration,  the  parameter space $\Lambda$ is equal to $\{0,1\}^3\backslash(0,0,0)$  so the underlying action set $\A$ corresponds to 7 possible binary (zero / non-zero) settings of  $\alpha, \beta,\eta$.  We  consider an adversarial immediate reward function $R$ that scores a given action  (and hence the underlying configuration $\lambda \in \Lambda$)   proportionally to the error rates of $g_t(\D_{t+1}^{\lambda})$; put differently, the display $\D_{t+1}^{\lambda}$ is  selected in order to challenge (the most) the current classifier $g_t$, leading to a better estimate of  $g_{t+1}$. With this RL-based  design,  better change detection performances are observed  as shown subsequently  in experiments.
\section{Experiments}\label{exp}
\indent {\bf Dataset and setting.}  We evaluate the accuracy of our RL-based interactive change detection algorithm using the Jefferson dataset.  The latter consists of $2,200$ non-overlapping ($30\times30$ RGB) patch pairs taken from bi-temporal GeoEye-1 satellite images of $2, 400 \times  1, 652$ pixels with a spatial resolution of 1.65m/pixel.  These patch pairs pave a large area from Jefferson (Alabama) in 2010 and in 2011.  These images show several damages caused by tornadoes (building destruction, debris on roads, etc) as well as no-changes including irrelevant ones (clouds, etc). In this dataset $2,161$ patch pairs correspond to negative data and only $39$ pairs to positive, so $<2\%$ of these data correspond to relevant changes and this makes their detection very challenging. In our experiments, half of the patch pairs are used for training and the remaining ones for testing.  We measure the accuracy of change detection using the equal error rate (EER); the latter is a balanced generalization error that evenly weights errors in  the positive and negative classes. Smaller EERs imply better performances.  \\ 
\begin{table}
 \centering 
 \resizebox{0.95\columnwidth}{!}{
 \begin{tabular}{c||cccccccccc|c}
Iter  & 1 &2 & 3& 4& 5& 6& 7& 8& 9 & 10  & AUC \\
Samp\%  & 1.45 &2.90 & 4.36& 5.81& 7.27& 8.72& 10.18& 11.63& 13.09 & 14.54& \\
 \hline
 \hline
  rep & 48.05 &  26.21 &  12.72 &   10.48 &    9.88 &   9.70&    8.52 &   8.85&   8.61&   8.82& 15.18 \\
  div&  48.05 &  31.24 &  23.45 &  30.41 &  44.81   &  24.12 &   13.22 &   17.02&    6.88. &   7.98 & 24.71 \\
 amb & 48.05 & 46.68 &   38.73 &   29.91 &   14.74 &   20.11 &   8.33 &   7.41 &   7.37 &   5.53 &22.68 \\ 
 rep+div & 48.05  &  26.21 &   33.35 &   25.10 &   21.55 &   11.71 &   2.84 &   1.65 &   1.59 &   1.43&  17.34 \\
 rep+amb &  48.05 &   26.21 &   12.62 &   10.81 &   9.82 &   9.70 &   8.53 &   9.23 &   8.60&   8.82& 15.23 \\ 
   div+amb &   48.05 &  41.69 &   28.82 &  23.08 &   23.41 &   23.42 &   19.82 &   13.10 &   8.16 &   6.97& 23.65 \\

   all (flat)  & 48.05 &  26.21 &  33.35 &   25.52 &   23.70 &  14.59 &   2.74 &   1.54&   1.67   & 1.48&  17.88 \\
   \hline \hline
      RL-based &   48.05 & 31.75&  10.36&   14.83&   13.36&   14.70&   1.06 & 1.06  & 1.10  &  \bf1.01 & \bf 13.72                                                                                          
\end{tabular}}
 \caption{This table shows an ablation study of our display model. Here rep, amb and div stand for representativity, ambiguity and diversity respectively. These results are shown for different iterations $t$ (Iter) and the underlying sampling rates (Samp)  defined as $(\sum_{k=0}^{t-1} |\D_k|/(|\X|/2))\times 100$. The AUC (Area Under Curve) corresponds to the average of EERs across iterations.}\label{tab1} 
 \end{table} 
 
\noindent {\bf Ablation study and impact of RL.} In the first set of experiments, we show an ablation study of our display model and thereby the impact of ambiguity, representativity and diversity  criteria when taken individually and combined. From these results, we observe the positive impact of diversity at the early iterations of active learning, while the impact of ambiguity comes later in order to further refine the learned change detection functions. However, none of the settings (rows) in table~\ref{tab1}  obtains the best performance through all the iterations of active learning. Considering these observed ablation performances, a better  setting of the $\alpha$, $\beta$ and $\eta$ should be cycle-dependent using reinforcement learning (as described in section~\ref{rlbased}), and as also corroborated through performances shown in table~\ref{tab1}. Indeed, it turns out that this adaptive setting outperforms  the other combinations (including ``all'',  also referred to as ``flat''), especially at the later iterations of change detection. \\

\begin{figure}[tbp]
\center
\includegraphics[width=0.69\linewidth]{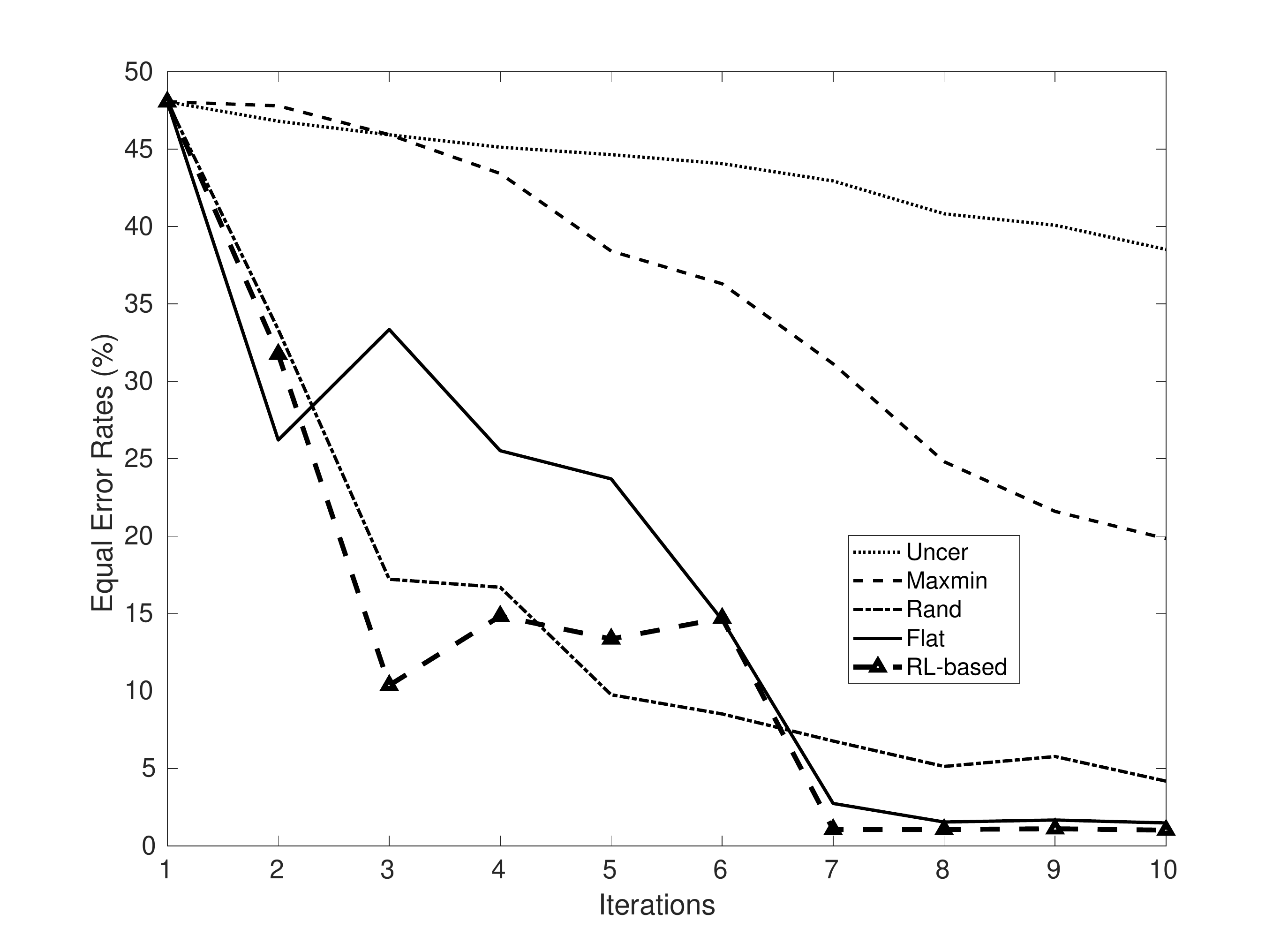}
 \caption{This figure shows a comparison of different sampling strategies w.r.t. different iterations (Iter) and the underlying sampling rates in table~\ref{tab1} (Samp). Here Uncer, Rand and Flat stand for uncertainty, random and all (rep+div+amb) sampling respectively. Note that fully-supervised learning achieves an EER of $0.94 \%$. See again section~\ref{exp} for more details.}\label{tab2}\end{figure}

\noindent {\bf Extra comparison.} Figure.~\ref{tab2} shows other comparisons of our RL-based display model w.r.t. different related display sampling techniques including {\it random, MaxMin and uncertainty}.  Random picks data from the unlabeled set whereas MaxMin greedily selects a sample $\x_i$  in ${\cal D}_{t+1}$ from the pool ${\X}\backslash \cup_{k=0}^t {\cal D}_k$ by maximizing its minimum distance w.r.t  $\cup_{k=0}^t {\cal D}_k$. We also compare our method w.r.t. uncertainty which consists in selecting samples in the display whose scores are the closest to zero (i.e., the most ambiguous). Finally, we also consider the fully supervised setting as an upper bound on performances; this configuration relies on the whole annotated training set and builds the learning model in one shot. \\ 

The EERs  in figure~\ref{tab2} show the  impact of the proposed RL-based display model against the related sampling strategies for different amounts of annotated data. The comparative methods are effective either at the early iterations of active learning (such as "random" which captures the diversity of data without being able to refine decision functions) or at the latest iterations (such as uncertainty which locally refines change detection functions but suffers from the lack of diversity). In contrast, our proposed RL-based design adapts the choice of these criteria as active learning cycles evolve, and thereby allows our interactive change detection to reach lower EERs and to overtake all the other strategies at the end of the iterative process.

\section{Conclusion} 
We introduce in this paper  a satellite image change detection algorithm based on active and reinforcement learning.  The strength of the proposed method resides in its ability to find and adapt display selection criteria to the active learning iterations,  thereby leading to more informative subsequent displays and more accurate decision functions.  Extensive experiments conducted on the challenging task of change detection shows the accuracy and the out-performance of the proposed interactive method w.r.t. the related work.
 \vfill\pagebreak

 {

}


\begin{thebibliography}{1}
\bibitem{ref4}  D. Brunner, G. Lemoine, and L. Bruzzone, Earthquake damage assessment of buildings using vhr optical and sar imagery, IEEE Trans. Geosc. Remote Sens., vol. 48, no. 5, pp. 2403--2420, 2010.

\bibitem{icassp2017b} M. Jiu and H. Sahbi. Deep kernel map networks for image annotation. IEEE International Conference on Acoustics, Speech and Signal Processing (ICASSP), 2016.


\bibitem{ref5} H. Gokon, J. Post, E. Stein, S. Martinis, A. Twele, M. Muck, C. Geiss, S. Koshimura, and M. Matsuoka, A method for detecting buildings destroyed by the 2011 tohoku earthquake and tsunami using multitemporal terrasar-x data, GRSL, vol. 12, no. 6, pp. 1277--1281, 2015.

\bibitem{ref7} J. Deng, K. Wang, Y. Deng, and G. Qi, PCA-based land-use change detection and analysis using multitemporal and multisensor satellite data, IJRS, vol. 29, no. 16, pp. 4823--4838, 2008.

\bibitem{sahbi00003} M. Jiu and H. Sahbi. "DHCN: Deep Hierarchical Context Networks For Image Annotation." ICASSP 2021-2021 IEEE International Conference on Acoustics, Speech and Signal Processing (ICASSP). IEEE, 2021.

\bibitem{ref9} R. Radke, S. Andra, O. Al-Kofahi, and B. Roysam, Image change detection algorithms: A systematic survey, IEEE Trans. on Im Proc, vol. 14, no. 3, pp. 294--307, 2005.

 \bibitem{ref11} S. Liu, L. Bruzzone, F. Bovolo, M. Zanetti, and P. Du, Sequential spectral change vector analysis for iteratively discovering and detecting multiple changes in hyperspectral images, TGRS, vol. 53, no. 8, pp. 4363--4378, 2015.

 \bibitem{sahbiICPR2020}
H. Sahbi. "Learning Connectivity with Graph Convolutional Networks." 2020 25th International Conference on Pattern Recognition (ICPR). IEEE, 2021.


\bibitem{ref13} G. Chen, G. J. Hay, L. M. Carvalho, and M. A. Wulder, Object-based change detection, IJRS, vol. 33, no. 14, pp. 4434--4457, 2012.


\bibitem{sahbi00009} M. Jiu and H. Sahbi. "Laplacian deep kernel learning for image annotation." 2016 IEEE International Conference on Acoustics, Speech and Signal Processing (ICASSP). IEEE, 2016.

\bibitem{ref14} J. Zhu, Q. Guo, D. Li, and T. C. Harmon, Reducing mis-registration and shadow effects on change detection in wetlands, Photogrammetric Engineering \& Remote Sensing, vol. 77, no. 4, pp. 325--334, 2011.

\bibitem{ref15} A. Fournier, P. Weiss, L. Blanc-Fraud, and G. Aubert, A contrast equalization procedure for change detection algorithms: applications to remotely sensed images of urban areas, In ICPR, 2008

\bibitem{Sahbi2011}
H.~Sahbi, J.-Y. Audibert, and R.~Keriven, ``Context-dependent kernels for  object classification,'' \emph{IEEE Transactions on Pattern Analysis and   Machine Intelligence}, vol.~33, pp. 699--708, 2011.

\bibitem{ref17} Carlotto, Detecting change in images with parallax, In Society of Photo-Optical Instrumentation Engineers, 2007 
\bibitem{ref20}  S. Leprince, S. Barbot, F. Ayoub, and J.-P. Avouac, Automatic and precise orthorectification, coregistration, and subpixel correlation of satellite images, application to ground deformation measurements,TGRS, vol. 45, no. 6, pp. 1529--1558, 2007.
  

\bibitem{ref21} Pollard, Comprehensive 3d change detection using volumetric appearance modeling, Phd, Brown University, 2009.

  \bibitem{sahbiiccv17}
  H. Sahbi. Coarse-to-fine deep kernel networks. Proceedings of the IEEE International Conference on Computer Vision, 1131-1139, 2017.
\bibitem{ref25}A. A. Nielsen, The regularized iteratively reweighted mad method for change detection in multi-and hyperspectral data, IEEE Transactions on Image processing, vol. 16, no. 2, pp. 463--478, 2007.


\bibitem{ref26} C. Wu, B. Du, and L. Zhang, Slow feature analysis for change detection in multispectral imagery, TGRS, vol. 52, no. 5, pp. 2858--2874, 2014

\bibitem{ref27} N. Bourdis, D. Marraud, and H. Sahbi, Spatio-temporal interaction for aerial video change detection, in IGARSS, 2012, pp. 2253--2256 


\bibitem{ref28} J. Im, J. Jensen, and J. Tullis, Object-based change detection using correlation image analysis and image segmentation, International Journal of Remote Sensing, vol. 29, no. 2, pp. 399--423, 2008.

 \bibitem{reff45} Vinyals et al., Matching networks for one shot learning. 2016. 

\bibitem{reff1}  Dasgupta, Analysis of a greedy active learning strategy, [Link.]

 \bibitem{Jiu2015}
M.~Jiu and H.~Sahbi, ``Semi supervised deep kernel design for image annotation,'' in \emph{ICASSP}, 2015.


 \bibitem{reff2} Settles, Active Learning: Synthesis Lectures on Artificial Intelligence and Machine Learning, [Link.]



\bibitem{reff13}  Tianxu et al., An Active Learning Approach with Uncertainty, Representativeness, and Diversity,


\bibitem{ref00004}   H. Sahbi and X. Li. "Context-based support vector machines for interconnected image annotation." Asian Conference on Computer Vision. Springer, Berlin, Heidelberg, 2010.


\bibitem{reff16}  Joshi et al., Multi-class active learning for image classification. 2009. 
 
\bibitem{reff15} Settles \& Craven. An analysis of active learning strategies for sequence labeling tasks. 2008.

\bibitem{reff53} Houlsby et al., Bayesian active learning for classification and preference learning. 2011. 

\bibitem{reff12}  Campbell \& Broderick, Automated scalable Bayesian inference via Hilbert coresets. 2019.

\bibitem{reff74} Gal et al., Deep bayesian active learning with image data. 2017 

\bibitem{Jiu2017}
  M.~Jiu and H.~Sahbi, ``Nonlinear deep kernel learning for image annotation,'' \emph{IEEE  Transactions on Image Processing}, vol. 26(4), 2017.


\bibitem{reff58} Pang et al., Meta-Learning Transferable Active Learning Policies by Deep Reinforcement Learning
  
\bibitem{refff1}  F.R. Bach. Active learning for misspecified generalized linear models. Advances in Neural Information Processing Systems (NIPS), 19, 2006.
\bibitem{refff2}  A. Kolesnikov, X. Zhai, L. Beyer. Revisiting Self-Supervised Visual Representation Learning  Proceedings of the IEEE/CVF Conference on Computer Vision and Pattern Recognition (CVPR), 2019, pp. 1920-1929

\bibitem{refff33333} H. Sahbi, S. Deschamps, A. Stoian. Frugal Learning for Interactive Satellite Image Change Detection. IEEE International Geoscience and Remote Sensing Symposium IGARSS, 2021.

\bibitem{refff333334} 
  Sutton, Richard S., and Andrew G. Barto. Reinforcement learning: An introduction. MIT press, 2018.

\bibitem{refff333335} 
  Jin, C., Allen-Zhu, Z., Bubeck, S., \& Jordan, M. I. (2018). Is Q-learning provably efficient?. arXiv preprint arXiv:1807.03765.


\bibitem{sahbi00007} P. Vo and H. Sahbi. "Transductive kernel map learning and its application to image annotation." BMVC. 2012.

\bibitem{aaaa0000}  Chen, Hao, and Zhenwei Shi. ”A spatial-temporal attention-based method and a new dataset for remote sensing image change detection.” Remote Sensing 12.10 (2020): 1662.


\bibitem{aaaa0001} Khan, Salman H., et al. ”Learning deep structured network for weakly supervised change detection.” arXiv preprint arXiv:1606.02009 (2016).

\bibitem{ref00007777} F. Fleuret and H. Sahbi. "Scale-invariance of support vector machines based on the triangular kernel." 3rd International Workshop on Statistical and Computational Theories of Vision. 2003.
\bibitem{ref00005} N. Boujemaa, F. Fleuret, V. Gouet, H. Sahbi (2004, January). Visual content extraction for automatic semantic annotation of video news. In the proceedings of the SPIE Conference, San Jose, CA (Vol. 6).

\bibitem{aaaa0002}Shi, Wenzhong, et al. ”Change detection based on artificial intelligence: State-of-the-art and challenges.” Remote Sensing 12.10 (2020):1688

\bibitem{ref00007} H. Sahbi, L. Ballan, G. Serra, A. DelBimbo (2012). Context-dependent logo matching and recognition. IEEE Transactions on Image Processing, 22(3), 1018-1031.

\bibitem{aaaa0003}Acito, Nicola, et al. ”Introductory view of anomalous change detection in hyperspectral images within a theoretical Gaussian framework.” IEEE Aerospace and Electronic Systems Magazine 32.7 (2017): 2-27.

\bibitem{aaaa0004}Vizilter, Yu V., et al. ”CHANGE DETECTION VIA MORPHOLOGICAL COMPARATIVE FILTERS.” ISPRS Annals of Photogrammetry, Remote Sensing and Spatial Information Sciences 3.3 (2016).

\bibitem{Sahbi20088888} M. Ferecatu and H. Sahbi. "TELECOMParisTech at ImageClefphoto 2008: Bi-Modal Text and Image Retrieval with Diversity Enhancement." CLEF (Working Notes). 2008.

\bibitem{aaaa0005}Han, Pengcheng, et al. ”Aerial image change detection using dual regions of interest networks.” Neurocomputing 349 (2019): 190-201.


\bibitem{aaaa0006} Mesquita, Daniel B., et al. ”Fully convolutional siamese autoencoder for change detection in UAV aerial images.” IEEE Geoscience and Remote Sensing Letters 17.8 (2019): 1455-1459.

\bibitem{sahbikpca06} H. Sahbi.  Kernel PCA for similarity invariant shape recognition. Neurocomputing 70 (16-18), 3034-3045

\bibitem{aaaa0007} Ignatiev, Vladimir, et al. ”Targeted change detection in remote sensing images.” Eleventh International Conference on Machine Vision (ICMV 2018). Vol. 11041. International Society for Optics and Photonics, 2019.

\bibitem{aaaa0008} Chianucci, Dan, and Andreas Savakis. ”Unsupervised change detection using spatial transformer networks.” 2016 IEEE Western New York Image and Signal Processing Workshop (WNYISPW). IEEE, 2016

\bibitem{aaaa0009} Zerrouki, Nabil, Fouzi Harrou, and Ying Sun. ”Statistical monitoring of changes to land cover.” IEEE Geoscience and Remote Sensing Letters 15.6 (2018): 927-931.

\bibitem{sahbi00001}  H. Sahbi, "Lightweight Connectivity In Graph Convolutional Networks For Skeleton-Based Recognition." 2021 IEEE International Conference on Image Processing (ICIP). IEEE, 2021.

\bibitem{aaaa0010} Bu, Shuhui, et al. ”Mask-CDNet: A mask based pixel change detection network.” Neurocomputing 378 (2020): 166-178.

\bibitem{aaaa0011} Konstantinidis, Dimitrios. ”Building detection for monitoring of urban changes.” (2017).
  
\bibitem{aaaa0012} Khan, Salman H., et al. ”Weakly supervised change detection in a pair of images.” arXiv preprint arXiv:1606.02009 (2016).

\bibitem{sahbi00005} H. Sahbi. "Kernel-based Graph Convolutional Networks." 2020 25th International Conference on Pattern Recognition (ICPR). IEEE, 2021.

\bibitem{aaaa0013} Kolos, Maria, et al. ”Procedural synthesis of remote sensing images for robust change detection with neural networks.” International Symposium on Neural Networks. Springer, Cham, 2019.

\bibitem{thiemert2005applying} S.~Thiemert, H.~Sahbi, and M.~Steinebach, ``Applying interest operators in  semi-fragile video watermarking,'' in \emph{Security, Steganography, and   Watermarking of Multimedia Contents VII}, vol. 5681.\hskip 1em plus 0.5em   minus 0.4em\relax International Society for Optics and Photonics, 2005, pp.  353--363.

\bibitem{aaaa0014} Shi, Qian, et al. ”A Deeply Supervised Attention Metric-Based Network and an Open Aerial Image Dataset for Remote Sensing Change Detection.” IEEE Transactions on Geoscience and Remote Sensing (2021).

\bibitem{aaaa0015} Tian, Shiqi, et al. ”Hi-UCD: A large-scale dataset for urban semantic change detection in remote sensing imagery.” arXiv preprint arXiv:2011.03247 (2020).

 \bibitem{sahbi00004} H. Sahbi. "Learning laplacians in chebyshev graph convolutional networks." Proceedings of the IEEE/CVF International Conference on Computer Vision. 2021.

\bibitem{aaaa0016} Hytla, Patrick C. Multi-Ratio Fusion Change Detection Framework with Adaptive Statistical Thresholding. Diss. University of Dayton, 2016.

\bibitem{aaaa0017} Andermatt, Philipp, and Radu Timofte. ”A weakly supervised convolutional network for change segmentation and classification.” arXiv preprint arXiv:2011.03577 (2020).

\bibitem{sahbi00006} A. Mazari and H. Sahbi. "MLGCN: Multi-Laplacian graph convolutional networks for human action recognition." The British Machine Vision Conference (BMVC). 2019.

\bibitem{aaaa0018} Hamaguchi, Ryuhei, et al. ”Epipolar-Guided Deep Object Matching for Scene Change Detection.” arXiv preprint arXiv:2007.15540 (2020).

\bibitem{aaaa0019} Li, Suicheng, et al. ”Change detection in images using shape-aware siamese convolutional network.” Engineering Applications of Artificial Intelligence 94 (2020): 103819.

\bibitem{aaaa0020} Mroueh, Fatima, Ihab Sbeity, and Mohamad Chaitou. ”Building Change Detection in Aerial Images.” BDCSIntell. 2019.

\bibitem{Sahbi2015} H.~Sahbi, ``Imageclef annotation with explicit context-aware kernel maps,''   \emph{International Journal of Multimedia Information Retrieval}, pp.  113--128, 2015.

\bibitem{aaaa0021} Resta, Salvatore. Anomalous change detection in multi-temporal hyperspectral images. 2012.

\bibitem{ref00004555} S. Tollari, P. Mulhem, M. Ferecatu, H. Glotin, M. Detyniecki, P. Gallinari, H. Sahbi and Z-Q. Zhao. "A comparative study of diversity methods for hybrid text and image retrieval approaches." In Workshop of the Cross-Language Evaluation Forum for European Languages, pp. 585-592. Springer, Berlin, Heidelberg, 2008.

\bibitem{aaaa0022} Lee, Haeyun, et al. ”Local Similarity Siamese Network for Urban Land Change Detection on Remote Sensing Images.” IEEE Journal of Selected Topics in Applied Earth Observations and Remote Sensing 14 (2021): 4139-4149.

 \bibitem{sahbiicpr2002} H. Sahbi and N. Boujemaa. "Coarse-to-fine support vector classifiers for face detection." Object recognition supported by user interaction for service robots. Vol. 3. IEEE, 2002.

\bibitem{aaaa0023} Shen, Li, et al. ”S2Looking: A Satellite Side-Looking Dataset for Building Change Detection.” arXiv preprint arXiv:2107.09244 (2021).
  
\bibitem{aaaa0024} Zhang, Lin, et al. ”Object-level change detection with a dual correlation attention-guided detector.” ISPRS Journal of Photogrammetry and Remote Sensing 177 (2021): 147-160.

\bibitem{aaaa0025} Javadi, Mohammad Saleh, Mattias Dahl, and Mats Pettersson. ”Change detection in aerial images using a Kendall’s TAU distance pattern correlation.” 2016 6th European Workshop on Visual Information Processing (EUVIP). IEEE, 2016.

\bibitem{sahbithesis} H. Sahbi, Coarse-to-fine support vector machines for hierarchical face detection. Diss. PhD thesis, Versailles University, 2003.

\bibitem{aaaa0026} Zhai, Xiaohua, et al. ”S4l: Self-supervised semi-supervised learning.” Proceedings of the IEEE/CVF International Conference on Computer Vision. 2019.
  
\bibitem{aaaa0027} Tung, Hsiao-Yu Fish, et al. ”Self-supervised learning of motion capture.” arXiv preprint arXiv:1712.01337 (2017).

\bibitem{sahbijstars17} Q. Oliveau, H. Sahbi. Learning attribute representations for remote sensing ship category classification.  IEEE Journal of Selected Topics in Applied Earth Observations and Remote Sensing, 2017. 

\bibitem{aaaa0028} Misra, Ishan, and Laurens van der Maaten. ”Self-supervised learning of pretext-invariant representations.” Proceedings of the IEEE/CVF Conference on Computer Vision and Pattern Recognition. 2020.

\bibitem{aaaa0029}  Hendrycks, Dan, et al. ”Using self-supervised learning can improve model robustness and uncertainty.” arXiv preprint arXiv:1906.12340 (2019).

\bibitem{aaaa0030}  Lan, Zhenzhong, et al. ”Albert: A lite bert for self-supervised learning of language representations.” arXiv preprint arXiv

\bibitem{sahbi00008} H. Sahbi and N. Boujemaa. "From coarse to fine skin and face detection." Proceedings of the eighth ACM international conference on Multimedia. 2000.

\bibitem{aaaa0031} Sermanet, Pierre, et al. ”Time-contrastive networks: Self-supervised learning from video.” 2018 IEEE international conference on robotics and automation (ICRA). IEEE, 2018.

\bibitem{aaaa0032} Liu, Xiao, et al. ”Self-supervised learning: Generative or contrastive.” IEEE Transactions on Knowledge and Data Engineering (2021).

  
\bibitem{aaaa0033} Zhou, Zhi-Hua. ”A brief introduction to weakly supervised learning.” National science review 5.1 (2018): 44-53.

\bibitem{ref000010}  H. Sahbi. "CNRS-TELECOM ParisTech at ImageCLEF 2013 Scalable Concept Image Annotation Task: Winning Annotations with Context Dependent SVMs." CLEF (Working Notes). 2013.

\bibitem{aaaa0034} Li, Yu-Feng, Lan-Zhe Guo, and Zhi-Hua Zhou. ”Towards safe weakly supervised learning.” IEEE transactions on pattern analysis and machine intelligence 43.1 (2019): 334-346.

\bibitem{ref00001244}  T. Stefan, H. Sahbi and M. Steinebach. "Using entropy for image and video authentication watermarks." Security, Steganography, and Watermarking of Multimedia Contents VIII. Vol. 6072. SPIE, 2006.

\bibitem{aaaa0035} Guo, Sheng, et al. ”Curriculumnet: Weakly supervised learning from large-scale web images.” Proceedings of the European Conference on Computer Vision (ECCV). 2018.

\bibitem{aaaa0036} Artzi, Yoav, and Luke Zettlemoyer. ”Weakly supervised learning of semantic parsers for mapping instructions to actions.” Transactions of the Association for Computational Linguistics 1 (2013): 49-62.

\bibitem{ref000012} H. Sahbi, J-Y. Audibert, R. Keriven. "Graph-cut transducers for relevance feedback in content based image retrieval." 2007 IEEE 11th International Conference on Computer Vision. IEEE, 2007.

\bibitem{aaaa0037} Yao, Xiwen, et al. ”Semantic annotation of high-resolution satellite images via weakly supervised learning.” IEEE Transactions on Geoscience and Remote Sensing 54.6 (2016): 3660-3671.

\bibitem{aaaa0038} Crandall, David J., and Daniel P. Huttenlocher. ”Weakly supervised learning of part-based spatial models for visual object recognition.” European conference on computer vision. Springer, Berlin, Heidelberg, 2006.
\bibitem{sahbijmlr06} H. Sahbi, D. Geman. A hierarchy of support vector machines for pattern detection. Journal of Machine Learning Research 7.Oct (2006): 2087-2123.

\bibitem{aaaa0039} Wang, Wei, et al. ”A survey of zero-shot learning: Settings, methods, and applications.” ACM Transactions on Intelligent Systems and Technology (TIST) 10.2 (2019): 1-37.


\bibitem{aaaa0040} Kodirov, Elyor, Tao Xiang, and Shaogang Gong. ”Semantic autoencoder for zero-shot learning.” Proceedings of the IEEE conference on computer vision and pattern recognition. 2017.

\bibitem{aaaa0041} Changpinyo, Soravit, et al. ”Synthesized classifiers for zero-shot learning.” Proceedings of the IEEE conference on computer vision and pattern recognition. 2016.

\bibitem{aaaa0042} Palatucci, Mark M., et al. ”Zero-shot learning with semantic output codes.” (2009).

\bibitem{ref000011}  H. Sahbi. "A particular Gaussian mixture model for clustering and its application to image retrieval." Soft Computing 12.7 (2008): 667-676.

\bibitem{aaaa0043} Zhang, Li, Tao Xiang, and Shaogang Gong. ”Learning a deep embedding model for zero-shot learning.” Proceedings of the IEEE conference on computer vision and pattern recognition. 2017

\bibitem{aaaa0044} Zhang, Ziming, and Venkatesh Saligrama. ”Zero-shot learning via semantic similarity embedding.” Proceedings of the IEEE international conference on computer vision. 2015.

\bibitem{aaaa0045} Verma, Vinay Kumar, et al. ”Generalized zero-shot learning via synthesized examples.” Proceedings of the IEEE conference on computer vision and pattern recognition. 2018.

\bibitem{ref000013} H. Sahbi, P. Etyngier, J-Y. Audibert, R. Keriven (2008, June). Manifold learning using robust graph laplacian for interactive image search. In 2008 IEEE Conference on Computer Vision and Pattern Recognition (pp. 1-8). IEEE.

\bibitem{aaaa0046} Zhu, Yizhe, et al. ”A generative adversarial approach for zero-shot learning from noisy texts.” Proceedings of the IEEE conference on computer vision and pattern recognition. 2018.


\bibitem{aaaa0047} Annadani, Yashas, and Soma Biswas. ”Preserving semantic relations for zero-shot learning.” Proceedings of the IEEE Conference on Computer Vision and Pattern Recognition. 2018.


\bibitem{aaaa0048} Noroozi, Mehdi, et al. ”Boosting self-supervised learning via knowledge transfer.” Proceedings of the IEEE Conference on Computer Vision and Pattern Recognition. 2018.

\bibitem{ref000014}   T. Napoléon and H. Sahbi. "From 2D silhouettes to 3D object retrieval: contributions and benchmarking." EURASIP Journal on Image and Video Processing 2010 (2010): 1-17.

\bibitem{aaaa0049} Medlock, Ben, and Ted Briscoe. ”Weakly supervised learning for hedge classification in scientific literature.” Proceedings of the 45th annual meeting of the association of computational linguistics. 2007.

\bibitem{aaaa0050} Maltezos, Evangelos, et al. ”Building Change Detection using Semantic Segmentation on Analogue Aerial Photos.” Proceedings of the
FIG Congress, Istanbul, Turkey. 2018.

\bibitem{ref00003}   A. Dutta and H. Sahbi. "High order stochastic graphlet embedding for graph-based pattern recognition." arXiv preprint arXiv:1702.00156 (2017).

\bibitem{aaaa0051} Xie, Guo-Sen, et al. ”Attentive region embedding network for zero-shot learning.” Proceedings of the IEEE/CVF Conference on Computer Vision and Pattern Recognition. 2019.

\end{thebibliography}
\end{document}